\documentclass{article}

\usepackage[final]{neurips_2023}
\usepackage{amsmath}
\usepackage{wrapfig}
\usepackage{algorithm}
\usepackage{algpseudocode}
\usepackage{graphicx}

\usepackage[utf8]{inputenc} 
\usepackage[T1]{fontenc}    
\usepackage{hyperref}       
\usepackage{url}            
\usepackage{booktabs}       
\usepackage{amsfonts}       
\usepackage{nicefrac}       
\usepackage{microtype}      
\usepackage{xcolor}         
\usepackage{caption}

\definecolor{mydarkblue}{rgb}{0,0.08,0.45}
\hypersetup{
  pdftitle={},
  pdfsubject={Proceedings of the International Conference on Machine Learning 2025},
  pdfkeywords={},
  pdfborder={0 0 0},
  pdfpagemode=UseNone,
  colorlinks=true,
  linkcolor=mydarkblue,
  citecolor=mydarkblue,
  filecolor=mydarkblue,
  urlcolor=mydarkblue
}

\title{SPIRe: Boosting LLM Inference Throughput with Speculative Decoding}

\author{
  Sanjit Neelam \quad Daniel Heinlein \quad Vaclav Cvicek \quad Akshay Mishra \quad Reiner Pope \\
  MatX \\
  \texttt{\{sanjit,daniel,vaclav,akshay,reiner\}@matx.com}
}

\begin{document}

\maketitle

\begin{abstract}
Speculative decoding (SD) has been shown to reduce the latency of autoregressive decoding (AD) by 2-3× for small batch sizes. However, increasing throughput and therefore reducing the cost per token requires decoding with large batch sizes. Recent work shows that SD can accelerate decoding with large batch sizes too if the context is sufficiently long and the draft model's KV cache is sparse. We introduce SPIRe, a draft model that combines static sparse attention, pruned initialization, and feedback memory to increase the modeled throughput of speculative decoding by over 100\% compared to speculation with a much smaller draft model and by over 35\% compared to the strong baseline of sparse self-speculation. Our approach is particularly effective when context lengths vary significantly across requests.
\end{abstract}

\section{Introduction}
Speculative decoding uses a cheap draft model to generate candidate tokens that are verified in parallel by an expensive target model \citep{leviathan2023, chen2023}. Accepted draft tokens are guaranteed to be samples from the \emph{target} model's distribution over the next token by a modified rejection sampling scheme. If draft tokens are generated quickly enough and have a high enough acceptance rate, speculative decoding improves both the latency and throughput of decoding by reducing the number of memory accesses needed to generate each token\footnote{The cost is that speculative decoding always requires performing more FLOPs than autoregressive decoding; the draft model performs forward passes, and the target model performs forward passes on not only the draft tokens that are accepted but also those that are rejected. Thus, speculative decoding only makes sense if autoregressive decoding is sufficiently memory-bound.}.

\begin{figure}[!htbp]
\centering
\includegraphics[width=\textwidth]{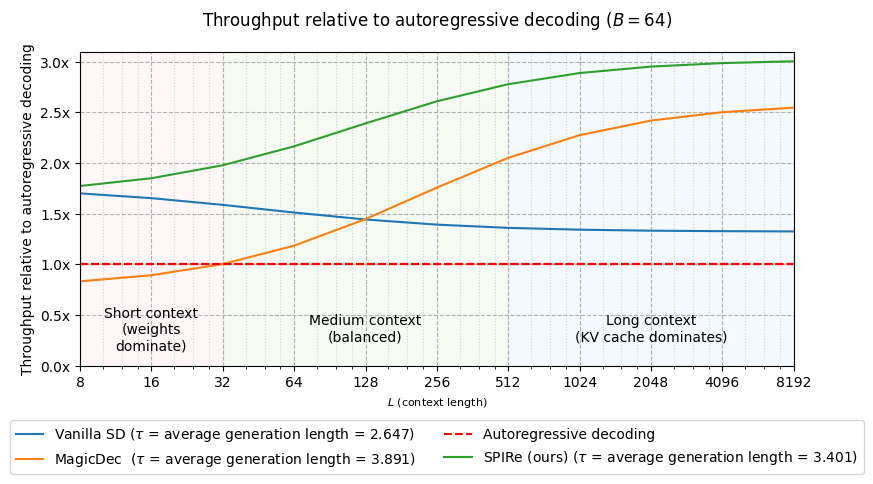}
\caption{Speedup of generating tokens in a batch of size $B=64$ given a context of length $L$. A \emph{round of speculation} is the generation and subsequent verification of each block of $k=4$ draft tokens, the \emph{average generation length} $\tau$ is the average number of tokens generated per round of speculation, and speedup is $\tau$ divided by how much longer a round of speculation takes compared to autoregressively decoding one token.}
\label{fig:1}
\end{figure}

Optimizing speculative decoding for higher throughput requires different strategies than optimizing for lower latency; the latter is well-documented in the literature \citep{xia2024, miao2024, yan2025}. For example, low latency can often be achieved with small batch sizes \citep{pope2022} and small draft models are preferred in this setting since fetching weights is the bottleneck. On the other hand, high throughput requires large batch sizes, which enables the use of larger draft models since in this regime loading the KV cache is the bottleneck \citep{chen2024}. See Figure \ref{fig:1} and Figure \ref{fig:4} for examples of how the speedup due to speculative decoding varies with draft model architecture, batch size, and context length according to our performance model \hyperref[sec:evaluation-metric]{below}.

Prior work has shown that the acceptance rate of draft tokens can be increased using the target model's intermediate activations and output logits \citep{du2024, zhou2024}. Since a target model's state can be cached during training at negligible cost, and since inference rather than training accounts for most of the cost of serving an LLM, the investment in training a draft model aligned to a particular target model may be worthwhile. Indeed, we show in \hyperref[sec:cost-analysis]{Cost Analysis} that this is the case for SPIRe under conservative assumptions.

Our work, along with \citet{chen2024}, demonstrates that speculative decoding is an effective technique for reducing the cost of decoding, and our work enables more efficient exploration of the design space of draft models based on the batch sizes and context lengths expected in production. Our main contributions are the following:

\begin{itemize}
\item An implementation-agnostic model for evaluating the throughput of speculative decoding with different draft models.
\item An efficient method for training a Feedback Transformer draft model.
\item A draft model, SPIRe, that increases the modeled throughput of speculative decoding by over 100\% compared to \hyperref[sec:evaluation]{vanilla speculative decoding} and by over 35\% compared to the strong baseline of MagicDec \citep{chen2024}.
\end{itemize}

\section{Evaluation Metric}\label{sec:evaluation-metric}
It is problematic to compare different methods of training the draft model using the throughput achieved by the resulting models in a production environment. Measured throughput depends on myriad implementation details such as use of specialized kernels \citep{dao2022}, how models are partitioned across chips \citep{pope2022}, how queries are batched \citep{daniel2023}, and which queries are used for evaluation \citep{zhou2024}. Any of these variables could be confounders when comparing different draft model architectures.

To derive a proxy for measured throughput, we first note that the throughput of speculative decoding is equal to the throughput of vanilla decoding multiplied by
\begin{align*}
    \begin{array}{c}
    \text{Throughput} \\
    \text{Multiplier}
    \end{array} &= \frac{\mathbb{E}[\text{\# Tokens Generated per Round of Speculation}]}{\frac{k \times t_\text{draft} + t_\text{verify}}{t_\text{target}}}
\end{align*}
where $k$ is the \emph{maximum speculation depth}, a \emph{round of speculation} is the generation and subsequent verification of each block of $k$ draft tokens, $t_\text{draft}$ and $t_\text{target}$ are the latencies of a draft and target model forward pass on a single token, and $t_\text{verify}$ is the latency of a target model forward pass on $k+1$ draft tokens in parallel. In the sequel, we refer to the \emph{average generation length} $\tau := \mathbb{E}[\text{\# Tokens Generated per Round of Speculation}]$ and the \emph{iteration time multiplier} $\Delta t := \frac{k \times t_\text{draft} + t_\text{verify}}{t_\text{target}}$, so that the throughput multiplier is simply $\tau / \Delta t$.

To obtain a more implementation-agnostic expression for $\Delta t$, we can break down the cost of a forward pass into the compute cost and the memory cost, which are the number of FLOPs and the number of memory accesses that need to be performed. The memory cost can be further broken down into the size of the LLM's weights and the size of its KV cache. By multiplying the memory cost in bytes with the hardware operational intensity (HOI) in FLOPs per byte, we can obtain an estimate of the memory cost in FLOP-equivalents \citep{matx2025} and compare the compute and memory costs on equal footing. Assuming that computation and memory accesses are maximally overlapped\footnote{Optimized inference stacks with e.g. pipeline parallelism approach this ideal. Throughput multipliers are higher than if we assumed that computation and memory accesses could not be overlapped, in which case we would have $\text{Decode Cost} = \text{Compute Cost} + \text{Memory Cost}$.}, we obtain the following expression for the cost of a forward pass.
\begin{align*}
    \text{Forward Cost} &= \max(\underbrace{\text{Compute Cost}}_{\text{FLOPs}}, \underbrace{\underbrace{(\text{Weight Cost} + \text{KV Cache Cost})}_{\text{Bytes}} \times \text{HOI}}_{\text{Memory Cost (FLOP-equivalents)}})
\end{align*}

Finally, by substituting $t_{\text{draft}}$, $t_{\text{verify}}$, and $t_{\text{target}}$ with their corresponding forward costs, we obtain the following implementation-agnostic\footnote{The inclusion of HOI in the throughput multiplier makes it hardware-dependent, but this only makes a difference in the compute-bound (bottom-left) region of Figure \ref{fig:4}. When decoding and verification are both memory-bound, HOI cancels out.} approximation of the throughput multiplier and use it as our evaluation metric.
\begin{align*}
    \begin{array}{c}
    \text{Throughput} \\
    \text{Multiplier}
    \end{array} \approx \frac{\mathbb{E}[\text{\# Tokens Generated per Round of Speculation}]}{\frac{k \times \max(C_\text{draft}, (N_\text{draft} + \text{KV}_\text{draft}) \times \text{HOI}) + \max((k + 1) \times C_\text{target}, (N_\text{target} + \text{KV}_\text{target}) \times \text{HOI})}{\max(C_\text{target}, (N_\text{target} + \text{KV}_\text{target}) \times \text{HOI})}}
\end{align*}
Here, $C_\text{draft}$ is the number of FLOPs performed during a forward pass through the draft model, $N_\text{draft}$ is the size of the draft model's weights in bytes, $\text{KV}_\text{draft}$ is the size of the draft model's KV cache in bytes, and similarly for $C_\text{target}$, $N_\text{target}$, and $\text{KV}_\text{target}$.

\section{SPIRe}
We combine the following techniques when training our draft model SPIRe in order to increase the throughput multiplier of speculative decoding. The name SPIRe stands for Sparse KV cache, Pruned Initialization, and Recurrent Feedback Transformer.

\subsection{Sliding window KV cache with attention sink}
In the large-batch long-context regime, the memory cost of loading the KV cache is the dominant cost of decoding, and reducing the size of the KV cache can reduce the iteration time multiplier $\Delta t$ by a larger factor than it reduces the average generation length $\tau$. Following MagicDec we use a StreamingLLM attention mask\footnote{\citet{xiao2024} use positions within the cache rather than those in the original text when adding positional information to tokens. We do the latter for SPIRe and the former for our implementation of MagicDec.} \citep{xiao2024}, during training and inference, to ensure the draft model's KV cache is sparse and that the size of the KV cache is constant with respect to the decoding sequence length. As shown in Figure \ref{fig:4}, this results in an increase in the throughput multiplier as the batch size increases, which is never the case for vanilla speculative decoding.

\subsection{Initialize by pruning the target model}
In our experiments with an eight-layer target model, we initialized the draft model using the embedding layer, the last two transformer blocks, and the unembedding layer of the trained target model. As shown in Figure \ref{fig:ablation}, initializing by pruning the target model produced a significantly higher value of $\tau$ than random initialization. Matching the embedding dimension of the target and draft models allowed us to further improve the quality of the draft model, as we'll explain below. Future work could explore more sophisticated pruning strategies, such as the one described by \citet{muralidharan2024}.

With short-to-medium contexts, MagicDec may underperform vanilla speculative decoding and even autoregressive decoding because the large increase in FLOPs outweighs the small decrease in KV cache loads. Given that context lengths vary considerably in production, we sought a draft model which accelerates decoding across a wide range of context lengths. Consequently, our draft model has $\frac{1}{4}$th as many parameters as the target model (ignoring embedding and unembedding parameters, which become negligible as the models are scaled up).

\subsection{Feedback Transformer and attending to target model activations}
Feedback memory \citep{fan2021} was proposed to increase the representation capacity of Transformers, creating Feedback Transformers with stronger performance at any given inference cost. They are significantly more expensive to train, however, because they inhibit parallelism over the context length. Fortunately, we find that this disadvantage can be almost entirely mitigated when training a draft model, since we can arrange for very short draft rollouts. Furthermore, the additional performance due to feedback memory has been shown to be even greater for shallow models such as ours, and a shallower draft model is preferable for its smaller forward pass latency $t_\text{draft}$.

The $\ell$-th attention sublayer of the Feedback Transformer computes
\begin{align*}
    \mathbf{z}_t^{\ell} = \text{Attention}(\mathbf{x}_t^{\ell}, \mathbf{m}_{<t})
\end{align*}
using memory vectors $\mathbf{m}_t$ instead of the standard $\mathbf{z}_t^{\ell} = \text{Attention}(\mathbf{x}_t^{\ell}, \mathbf{x}_{<t}^{\ell})$. The memory vectors $\mathbf{m}_t$ are defined as
\begin{align*}
    \mathbf{m}_t = \frac{\sum_{\ell=0}^{n_\text{layer}} \exp(w_\ell) \cdot \mathbf{x}_t^{\ell}}{\sum_{\ell=0}^{n_\text{layer}} \exp(w_\ell)}
\end{align*}
where $\mathbf{x}_t^{\ell}$ is the activation of the $t$-th token after the $\ell$-th layer, $(w_\ell) \in \mathbb{R}^{{n_\text{layer}}+1}$ are learnable weights, and layer 0 is the embedding layer. The dependence of each memory vector on all activations from the previous timestep does not increase the latency of decoding tokens autoregressively, but it does prevent training and prefill from being parallelized.

We require our Feedback Transformer draft model to be able to generate blocks of $k$ tokens given contexts of various lengths, where $k$ is the maximum speculation depth. Thus during training, rather than performing $S$ forward passes on a prefix of length $S$, we perform just $k$ forward passes: we forego computing the first $S - k$ memory vectors by substituting them with target model activations.

\begin{figure}[!htbp]
\centering
\includegraphics[width=\textwidth]{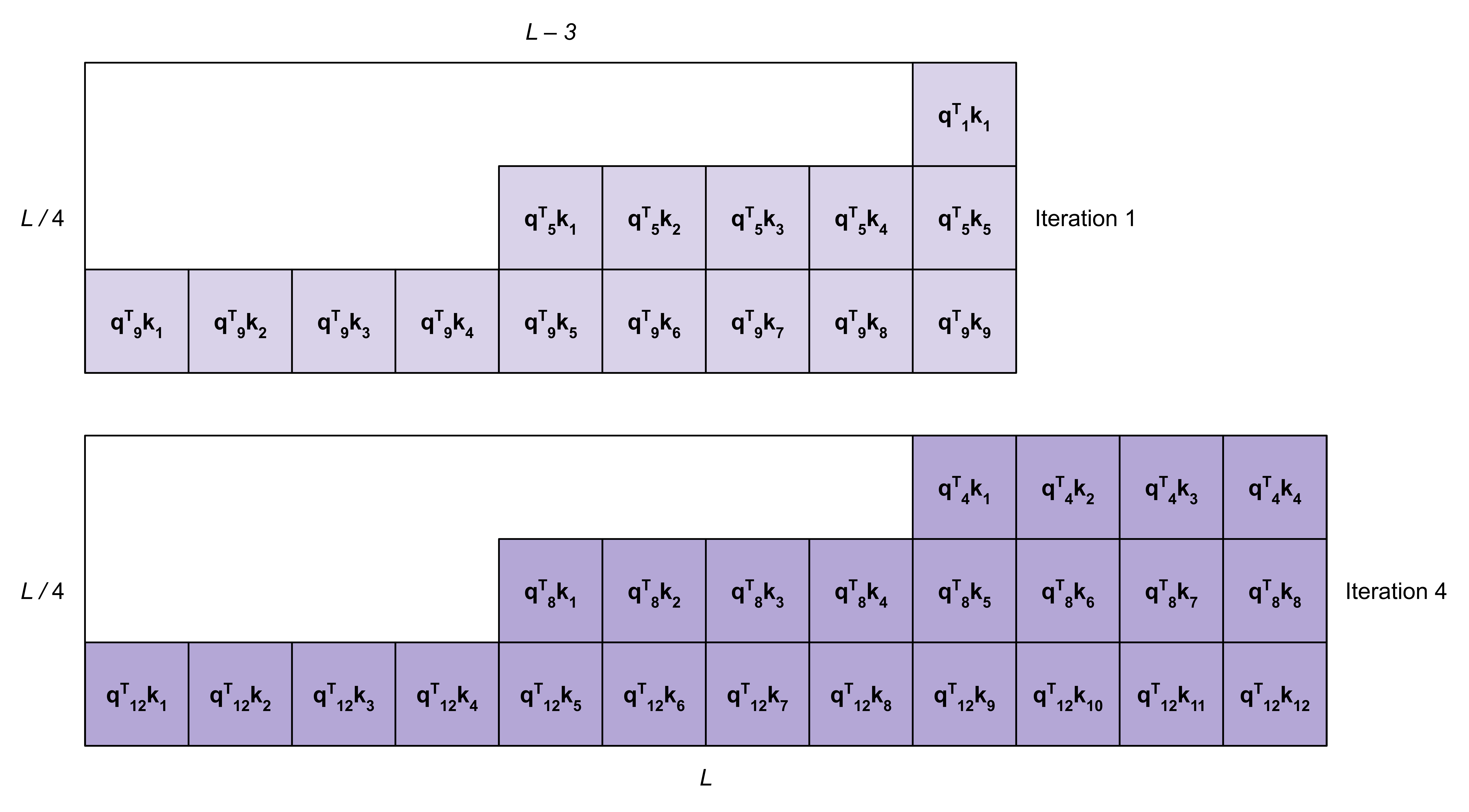}
\caption{The matrix $QK^\top$ in the 1st and 4th forward passes on a training sequence of length $L=12$ with a maximum speculation depth of $k=4$. In general we perform $k$ forward passes on every $k$-th prefix of a sequence, in a process similar to batched autoregressive decoding with teacher forcing.}
\label{fig:mask}
\end{figure}

Specifically, during training, our Feedback Transformer draft model autoregressively generates $k=4$ tokens following every $k$-th prefix of a sequence by performing $k$ forward passes. Unlike \citet{fan2021}, we share neither key and value projections nor memory vectors across layers: for $t = S - k + 1, \dots, S$, where $S \in \{k, 2k, \dots, L\}$ is the length of a prefix, the memory vectors $\mathbf{m}_t^i$ for the $i$-th layer are defined as
\begin{align*}
    \mathbf{m}_t^i = \frac{\sum_{\ell=0}^{n_\text{layer}} \exp(w_{i \ell}) \cdot \mathbf{x}_t^{\ell}}{\sum_{\ell=0}^{n_\text{layer}} \exp(w_{i \ell})}
\end{align*}
where $(w_{i \ell}) \in \mathbb{R}^{{n_\text{layer}} \times ({n_\text{layer}} + 1)}$ are learnable weights. For $t = 1, \dots, S - k$, we let $\mathbf{m}_t^i = \mathbf{y}_t^{i + 6 - 1}$, where $\mathbf{y}_t^{\ell}$ is the target model activation of the $t$-th token after the $\ell$-th layer. In other words, the weights of the first (second) layer of the draft model are initialized from the weights of the seventh (eighth) layer of the target model, and we let $\mathbf{m}_t^1 = \mathbf{y}_t^6$, $\mathbf{m}_t^2 = \mathbf{y}_t^7$ for $t = 1, \dots, S - k$.

If we did not match the embedding dimension of the draft and target model, we could have substituted past memory vectors with draft model embeddings, but this was worse than substituting with target model activations in our experiments. Our approach was also better than sharing memory vectors across layers and substituting past memory vectors with either draft model embeddings or post-final-layer target model activations.

\subsection{Distillation loss}
Early implementations of speculative decoding for LLMs \citep{leviathan2023, chen2023} optimize the draft model's parameters by minimizing the standard cross-entropy loss
\begin{align*}
   \text{HardTargetLoss} &= \mathbb{E}_{x_{<t} \sim D} \left[ - \sum_{i \in V} p_\text{data}(x_{<t})_i \log q(x_{<t})_i \right]
\end{align*}
where $x_{<t}$ is a prefix sampled from a dataset $D$, $p_\text{data}(x_{<t})$ is a one-hot vector corresponding to the token that follows $x_{<t}$, $q(x_{<t})$ is a dense vector corresponding to the draft model's probability distribution of the token that follows $x_{<t}$, and $V$ is the vocabulary size. To train a draft model to produce outputs that are more similar to the outputs of the target model, we minimize
\begin{align*}
   \text{MixedLoss}(\omega) = \omega \cdot \mathbb{E}_{x_{<t} \sim D} \left[ - \sum_{i \in V} p_\text{target}(x_{<t})_i \log q(x_{<t})_i \right] + (1 - \omega) \cdot (-\alpha)
\end{align*}
where $p_\text{target}(x_{<t})$ is a dense vector corresponding to the target model's probability distribution of the token that follows $x_{<t}$ and $\alpha$ is the expected acceptance probability \citep{leviathan2023}. Minimizing the first term minimizes the distillation loss \citep{hinton2015}, and minimizing the second term maximizes the average generation length $\tau$. As shown in Figure \ref{fig:ablation}, $\omega=0.5$ marginally outperformed $\omega=1$.

\section{Evaluation}\label{sec:evaluation}
We fix an 8-layer, 67-million body-parameter multi-head attention target model and vary the draft model, with all model architecture details provided in the Appendix. Each model with $N$ total-parameters is trained on $20 \times N$ tokens \citep{hoffmann2022} from the LongCrawl64 dataset \citep{buckman2024} using sequences of length 1024. We compare drafting tokens for the target model using the following draft models:

\begin{enumerate}
\item \textbf{Vanilla speculative decoding} \citep{chen2023}. This draft model has $\frac{1}{8}$th as many body parameters as the target model, and is trained by minimizing the standard cross-entropy loss. This is the most widely understood implementation of speculative decoding, and, as can be seen in Figures \ref{fig:1} and \ref{fig:4}, it is strong for short contexts and small batch sizes.
\item \textbf{MagicDec} \citep{chen2024}. This draft model is identical to the target model, but it sparsely accesses its KV cache through a StreamingLLM attention mask with a window size of 64 and a sink size of 1. This is an elegant implementation of speculative decoding, and is strong in the large-batch long-context regime.
\item \textbf{SPIRe} (ours). This draft model was described above in detail, and it has $\frac{1}{4}$th as many body parameters as the target model. During both training and inference, we use a StreamingLLm attention mask with a window size of 64 and a sink size of 1.
\end{enumerate}

We measure average generation lengths $\tau$ by generating $G=64$ tokens given $n=65536$ contexts from the validation split of the LongCrawl64 dataset, and we calculate iteration time multipliers $\Delta t$ for different batch sizes and context lengths using our performance model.

We use $\tau$'s corresponding to a medium context length of $L=512$ to calculate all throughput multipliers $\tau / \Delta t$ in Figures \ref{fig:1}, \ref{fig:4}, \ref{fig:ablation}, and \ref{fig:extrapolation}, by assuming that $\tau$ is the same when generating with shorter or longer contexts. Table \ref{tab:table} shows that the average generation length $\tau$ is similar for contexts of length $L \in \{256, 512, 960\}$, and our assumption is supported by e.g. Figure 5 of \citet{xiao2024}, which suggests that even with extremely long contexts, it suffices to attend to the most recent tokens.

\begin{table}[!htbp]
\centering
\caption{Average generation lengths $\tau$, with 95\% confidence intervals, for different draft models and context lengths. Using a maximum speculation depth of $k=4$, each round of speculation generates between 1 and 5 tokens, with between 0 and 4 tokens accepted before the first rejection.}
\label{tab:table}
\begin{tabular}{lccc}
\toprule
Context Length & Vanilla SD & MagicDec & SPIRe \\
\midrule
960 & 2.644 $\pm$ 0.004 & 3.793 $\pm$ 0.004 & 3.382 $\pm$ 0.004 \\
512 & 2.647 $\pm$ 0.004 & 3.891 $\pm$ 0.004 & 3.401 $\pm$ 0.004 \\
256 & 2.637 $\pm$ 0.004 & 4.005 $\pm$ 0.004 & 3.427 $\pm$ 0.004 \\
\bottomrule
\end{tabular}
\end{table}

\begin{figure}[!htbp]
\centering
\includegraphics[width=\textwidth]{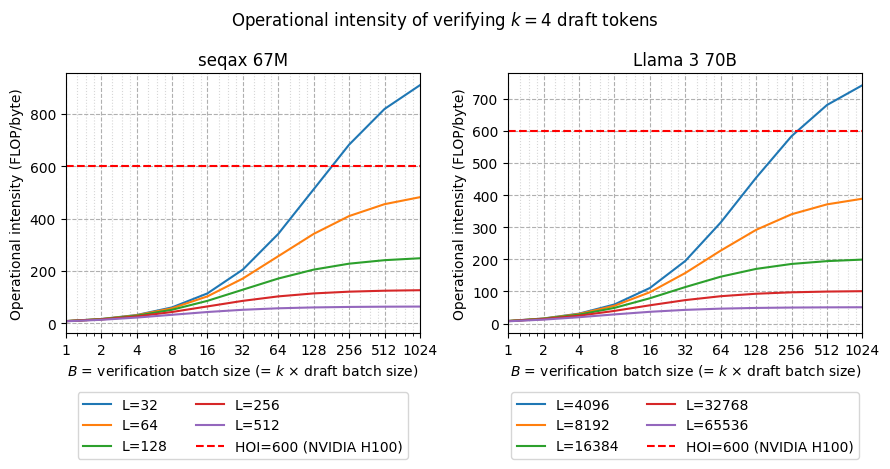}
\caption{With a sufficiently long context, verification may never be compute-bound even in the limit as batch size tends to infinity. For context lengths where verification is compute bound for some batch size $B_\text{crit}$, there's no benefit in terms of throughput (and some cost in terms of latency) to increasing batch size above $B_\text{crit}$.}
\label{fig:batch-size}
\end{figure}

We focus on a medium context of 512 tokens since we care about accelerating decoding across a range of context lengths. If a context of 128K is considered "long" for Llama 3 70B \citep{llama2024}, then a context of 64K may be considered "medium". As shown in Figure \ref{fig:batch-size}, 64K is four powers of two larger than the longest context $L_\text{crit}$ for which verification with Llama 3 70B is compute-bound at \emph{some} batch size. A context length of 512 is four powers of two larger than $L_\text{crit} = 32$ for our target model, so we consider a context of 512 to be of medium length.

\subsection{Results}
\begin{figure}[!htbp]
\centering
\includegraphics[width=0.85\textwidth]{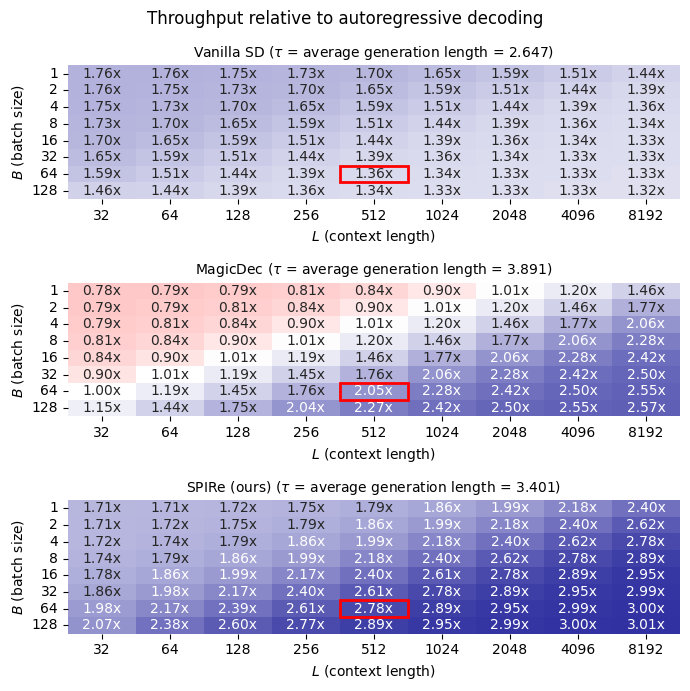}
\caption{Speedup of generating tokens in a batch of size $B$ given a context of length $L$. Using the highlighted values, we get that SPIRe increases the modeled throughput of speculative decoding by 100\% compared to vanilla speculative decoding and by 35\% compared to MagicDec.}
\label{fig:4}
\end{figure}

Our draft model SPIRe produces the largest increase in the modeled throughput of generating tokens for most batch sizes and context lengths, as shown in Figure \ref{fig:4}. SPIRe achieves a lower average generation length $\tau$ than MagicDec, but drafting with SPIRe nevertheless produces a larger speedup due to its lower cost of performing forward passes.

Figure \ref{fig:1} is derived from Figure \ref{fig:4} by focusing on a batch size of $B=64$. For medium contexts of around 512 tokens, Figure \ref{fig:batch-size} shows that the operational intensity of verification barely increases as we increase the verification batch size beyond $k \times 64 = 4 \times 64 = 256$, but it does increase significantly as the verification batch size is increased from 1 to 64.

\subsection{Ablations}\label{subsec:ablations}
\begin{figure}[!htbp]
\centering
\includegraphics[width=\textwidth]{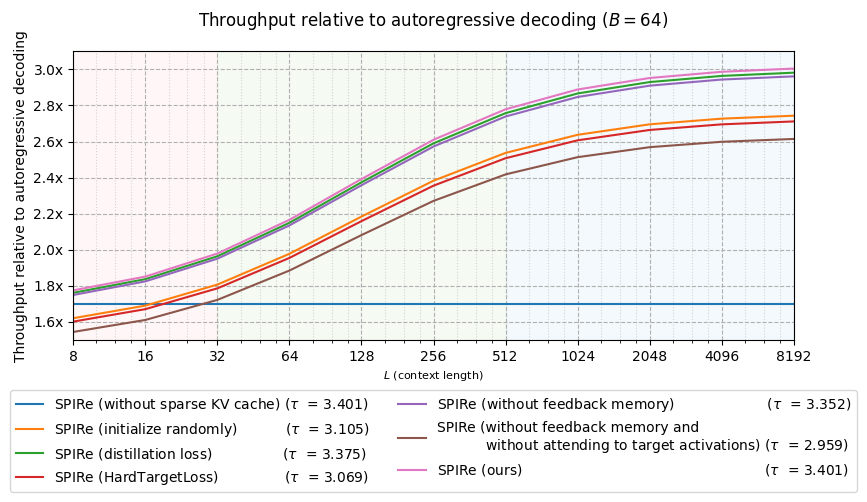}
\caption{Ablations of the techniques used to train our draft model SPIRe. In order of importance, the techniques are 1) sparse KV cache, 2) attending to target model activations, 3) distillation loss, 4) initializing by pruning the target model, 5) feedback memory, and 6) $\text{MixedLoss}$.}
\label{fig:ablation}
\end{figure}

See Figure \ref{fig:ablation} for ablations of the techniques used to train our draft model SPIRe. To ablate the sparse KV cache, we conservatively assume the same average generation length $\tau$ as SPIRe, and use our performance model to calculate new throughput multipliers. We ablate the other techniques by training new draft models and evaluating them.

The throughput multiplier for SPIRe (without sparse KV cache) is constant with respect to the context length $L$ in Figure \ref{fig:ablation} since here the memory cost of generating a token is higher than the compute cost, and since $k=4$ and $N_\text{target} = 4 \times N_\text{draft}$, $L$ cancels out in our expression for the iteration time multiplier $\Delta t$.

\subsection{Extrapolation}
\begin{figure}[!htbp]
\centering
\includegraphics[width=\textwidth]{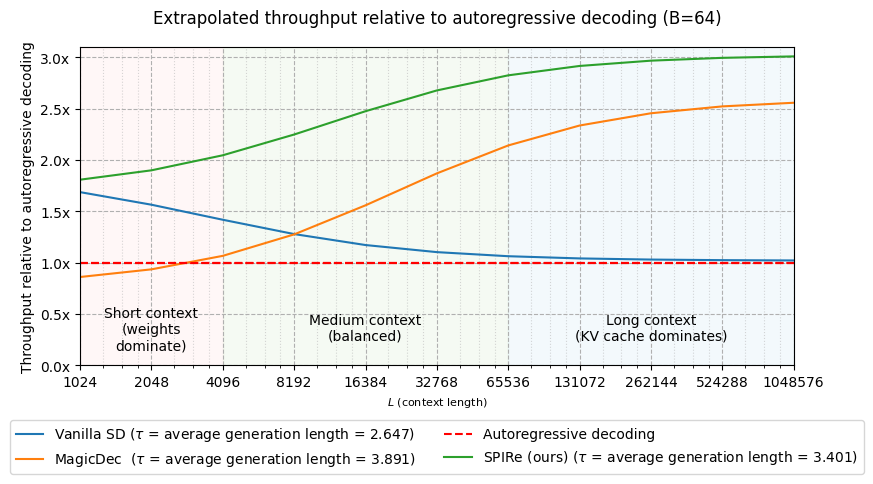}
\caption{Estimated speedup of generating tokens from Llama 3 70B in a batch of size $B$ given a context of length $L$. This figure is very similar, but not identical, to Figure \ref{fig:1}.}
\label{fig:extrapolation}
\end{figure}

What if we extrapolate the results above to drafting for Llama 3 70B while maintaining the ratio $N_\text{draft} / N_\text{target}$, by assuming the same values of the average generation length $\tau$ but recalculating iteration time multipliers using our performance model? Our draft model SPIRe produces the largest increase in the modeled throughput of generating tokens for most context lengths supported by LLM providers, as shown in Figure \ref{fig:extrapolation}.

\section{Cost Analysis}\label{sec:cost-analysis}
Each draft model costs a different amount in training FLOPs. We ignore FLOPs due to embedding and unembedding parameters, which become negligible as the draft model is scaled up. The cost to train our draft model SPIRe is around $\frac{1}{4}$th the cost to train the target model, since it has $\frac{1}{4}$th as many body parameters but achieves lower MFU during training.

Suppose that training the target model costs 1 unit of our compute budget, and that we expect to spend 10 units generating tokens for customers using autoregressive decoding. Using our draft model SPIRe for speculative decoding with $(B, L) = (64, 512)$ amounts to investing $0.25$ to train the draft model but saving $10 \times (1 - 1/2.78) = 6.40$ units in decoding cost, and drafting with MagicDec amounts to saving $10 \times (1 - 1/2.05) = 5.12$ units in decoding cost (the throughput multipliers 2.78 and 2.05 are highlighted in Figure \ref{fig:4}). Overall, SPIRe saves $6.15$ units, or over 20\% more than what MagicDec saves.

If we again extrapolate our results to drafting for Llama 3 70B with $(B, L) = (64, 65536)$, we find that drafting with SPIRe saves $10 \times (1 - 1/2.82) = 6.45$ units in decoding cost and drafting with MagicDec saves $10 \times (1 - 1/2.14) = 5.33$ units (the throughput multipliers 2.82 and 2.14 are highlighted in the Appendix). Overall, SPIRe saves $6.20$ units, or over 16\% more than what MagicDec saves. If there is significant downward variation in batch sizes and context lengths, then SPIRe outperforms MagicDec by a much greater amount in terms of decoding cost saved.

\section{Related Work}
\begin{itemize}
\item The only other work we're aware of which attempts to use speculative decoding to increase the throughput of decoding, rather than to minimize latency, is MagicDec \citep{chen2024}.
\item Continuous batching reduces the need for speculative decoding, but both techniques are complementary. In vLLM, speculative decoding is integrated with the system's continuous batching architecture, and studying their interaction further is an interesting direction for future work.
\item Our technique is orthogonal to any technique used to accelerate prefill. In practice, accelerating both prefill and decoding is important for reducing the cost per token generated.
\item We used the token verification algorithm in our experiments, but our results should hold when using the block verification \citep{sun2024} algorithm too.
\end{itemize}

\section{Conclusion}
Minimizing cost per token requires maximizing throughput. We demonstrated that draft model architecture significantly impacts throughput, with optimal choices depending on the batch sizes and context lengths expected in production. We proposed an implementation-agnostic performance model for evaluating the throughput of speculative decoding with different draft models, and we proposed a draft model SPIRe which outperforms strong baselines when decoding with large batch sizes and medium-to-long contexts. Future work can empirically validate our performance model, use a more principled long-context evaluation, and analyze the sensitivity of speedup with respect to the maximum speculation depth $k$.

\section{Contributions and Acknowledgments}
Sanjit Neelam proposed most aspects of the architecture of our draft model SPIRe. Feedback Transformer was suggested by Reiner Pope, and attending to target activations was co-developed by Sanjit Neelam and Reiner Pope. Explicitly maximizing the expected acceptance probability $\alpha$ was proposed by Akshay Mishra. The evaluation methodology was co-designed by Sanjit Neelam and Reiner Pope. All experiments were conducted by Sanjit Neelam, who also wrote all text and produced all figures. Daniel Heinlein and Vaclav Cvicek provided valuable feedback that improved the presentation of results.

We use \href{https://github.com/MatX-inc/seqax}{seqax}, our research-focused LLM codebase built on \href{https://github.com/jax-ml/jax}{JAX}, to perform all experiments. This work was supported by Cloud TPUs from Google's \href{https://sites.research.google/trc/about/}{TPU Research Cloud}.

\section*{Appendix}
See \url{https://github.com/MatX-inc/seqax/blob/SPIRe/spire_appendix.ipynb}.

\bibliographystyle{plainnat}
\bibliography{references}

\end{document}